\documentclass{article}
\pdfoutput=1
\usepackage{microtype}
\usepackage{graphicx}
\usepackage{balance}
\usepackage{subcaption}
\usepackage{float}
\usepackage{booktabs} % for professional tables
\graphicspath{{img/}} 

\usepackage{hyperref}

\usepackage[accepted]{icml2023}

\usepackage{amsmath}
\usepackage{amssymb}
\usepackage{mathtools}
\usepackage{amsthm}

\usepackage[capitalize,noabbrev]{cleveref}

\theoremstyle{plain}

\theoremstyle{definition}

\theoremstyle{remark}

\usepackage[textsize=tiny]{todonotes}

\icmltitlerunning{Terrain Classification Uncertainty for Space Exploration Robots from Proprioceptive Data}

\begin{document}

\twocolumn[
\icmltitle{Terrain Classification Enhanced with Uncertainty for Space Exploration Robots from Proprioceptive Data}

% \icmltitle{Terrain Classification Uncertainty for Space Exploration Robots \\ from Proprioceptive Data}



\icmlsetsymbol{equal}{*}

\begin{icmlauthorlist}
\icmlauthor{Mariela De Lucas Álvarez}{equal,DFKI}
\icmlauthor{Jichen Guo}{equal,Uni}
\icmlauthor{Raúl Domíguez}{DFKI}
\icmlauthor{Matias Valdenegro-Toro}{Gro}
\end{icmlauthorlist}

\icmlaffiliation{DFKI}{German Research Center for Artificial Intelligence}
\icmlaffiliation{Uni}{AG Robotik, University of Bremen}
\icmlaffiliation{Gro}{Bernoulli Institute, University of Groningen}

\icmlcorrespondingauthor{Mariela De Lucas Álvarez}{mariela.de\_lucas\_alvarez@dfki.de}
\icmlcorrespondingauthor{Jichen Guo}{jichen@uni-bremen.de}

\icmlkeywords{Terrain classification, uncertainty quantification, neural networks, space robotics}

\vskip 0.3in
]



\printAffiliationsAndNotice{\icmlEqualContribution} % otherwise use the standard text.

\begin{abstract}
Terrain Classification is an essential task in space exploration, where unpredictable environments are difficult to observe using only exteroceptive sensors such as vision. Implementing Neural Network classifiers can have high performance but can be deemed untrustworthy as they lack transparency, which makes them unreliable for taking high-stakes decisions during mission planning. We address this by proposing Neural Networks with Uncertainty Quantification in Terrain Classification. We enable our Neural Networks with Monte Carlo Dropout, DropConnect, and Flipout in time series-capable architectures using only proprioceptive data as input. We use Bayesian Optimization with Hyperband for efficient hyperparameter optimization to find optimal models for trustworthy terrain classification.

\end{abstract}

\section{Introduction}
\label{intro}
Terrain Classification (TC) is a common research problem in rover exploration, typically addressed through computer vision \cite{Liyanage2020,Wietrzykowski2014}. Given the many challenges that arise when deploying rovers for space exploration, integrating proprioceptive sensors to classify terrain is a practical solution.

We address TC for rovers using only proprioceptive data with the goal of enabling reliable extra-planetary missions. Environmental conditions that cause fluctuations in illumination or atmospherical conditions can impact the reliability of visual input. To mitigate this issue, we propose classifiers that are not reliant on visual input to maintain accuracy and robustness in the event of unforeseen environmental events. Specifically, we aim to use visual-independent or proprioceptive sensors such as Inertial Measurement Units (IMUs) and motor joint states to train a set of Neural Networks (NNs). The collected data contains terrain properties, which represent mobility characterizations from the navigation of the rover.

Neural networks still pose reliability concerns when being deployed in high-stakes applications \cite{li2023reliability}. We mitigate the issue of NN untrustworthiness by generating networks with Uncertainty Quantification (UQ) layers to Time Series Classification (TSC) architectures such as the Long Short-Term Memory Network (LSTM) \cite{Hochreiteretal1997lstm} and the Convolutional Neural Network (CNN) \cite{LeCun1998}. We also enhance state-of-the-art architectures which include the Fully Convolutional Network (FCN), a ResNet and an Attention Encoder \cite{Wang2017, Serra2018}.

We use Bayesian Optimization with Hyperband (BOHB) \cite{Falkner2018} for hyperparameter optimization to generate a set of high-accuracy CNN and LSTM candidates.  We then assess UQ and non-UQ networks in terms of predictive accuracy, Expected Calibration Error (ECE) \cite{Guo2017,Naeini2015} and use predictive entropy as a metric for UQ to determine the reliability of the classification.
% We assess the performance of the networks in terms of the overall predictive entropy

We quantitatively show the advantage of using UQ-NNs for reliably selecting good-performing candidates for a task where vision is traditionally used. We show that our networks are trustworthy for TC in critical settings with data that is difficult to interpret by simple observation.

% This paper is organized into the following sections. Section \ref{sec:rel_work} reviews relevant work in the field of Terrain Classification, its methodologies and the relevance of  TSC and UQ for this task. In Section \ref{sec:methods} we present our experimental methodology where we detail the NN architectures and how UQ has been integrated into the architectures. Finally, we present and discuss our results, and summarize our work in sections \ref{sec:res} and \ref{sec:conclusion}.

\section{Related Work}
\label{sec:rel_work}

\begin{figure}[t]
    \centering
    \includegraphics[width=.45\textwidth]{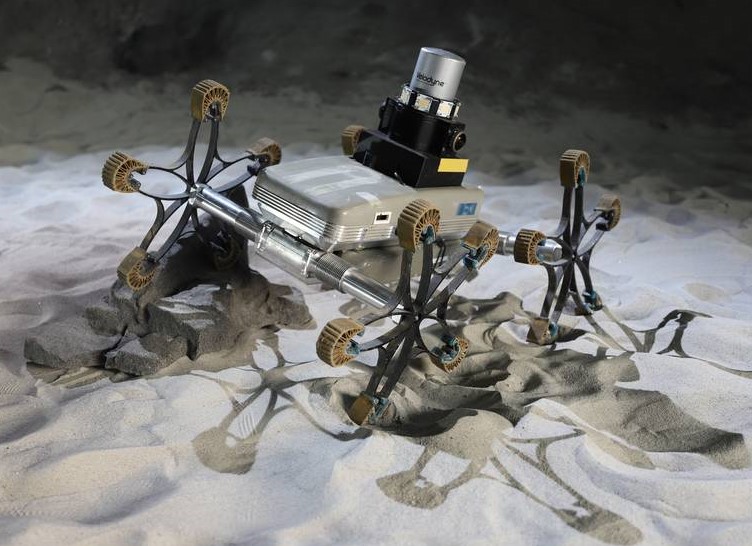}
    \caption{The AsguardIV is a hybrid leg-wheel robot designed at the DFKI to allow navigation in unstructured environments. Its rimless wheels are simpler, more energy-efficient, and more reliable than articulated legs, adapting effectively to obstacles and uneven terrain.}
    \label{fig:asguard}
\end{figure}

Identifying unknown terrain types has always been one fundamental challenge for mobile robot navigation. Existing research on Terrain Classification primarily focuses on visual or exteroceptive sensors to classify terrain by using navigation cameras \cite{Helmick2009, Rothrock2016}, RGB-D cameras \cite{Kozlowski2018}, hyperspectral cameras \cite{winkens2017,li2020} or LADAR \cite{Vandapel2004}. Although the visual features of different terrains provide important information to estimate the ground type nearby, some constraints for Terrain Classification with images include lighting and atmospherical conditions, such as inadequate lighting for a hyperspectral camera and clouds of gas, smoke, or sand when using a LADAR or standard cameras. 

Conversely, proprioceptive sensors provide information about the internal state of the robot and are therefore less susceptible to external factors that could compromise the acquisition of data. Existing work addresses TC by using proprioceptive sensors such as IMUs and torque sensors with traditional Machine Learning methods showing promising results \cite{Ojeda2006,Dimastrogiovanni2020,Vulpi2020,Dimastrogiovanni2021,Ugenti2022}.

% Therefore, more and more visual-independent sensors have been used and investigated in estimating terrains, which contain: vibration sensor \cite{Brooks2005}, motor current and voltage sensor \cite{Ojeda2006}, Inertial Measurement Unit (IMU) sensor \cite{Jitpakdee2008}, ultrasonic range sensor \cite{Ojeda2006}, force-torque sensor \cite{Dimastrogiovanni2021}. In the work \cite{Vulpi2020}, Vulpi et.al made use of data from motion states and wheel force and torque sensors, attached to a four-wheeled mobile robot named SherpaTT \cite{Cordes2017}, to identify 3 types of terrains: sandy, gravel and paved ground. The authors also compared the classification performance between Support Vector Machine(SVM) and CNN models under different sensor modalities, in which CNN models showed better results in all cases. We follow the research approach from \cite{Ugenti2022} which focuses exclusively on proprioceptive sensors using CNNs.

% Among the most successful DL methods used for TC are the LSTM and the CNN. The LSTMs are a type of Recurrent Neural Networks (RNN) \cite{Rumelhart1986rnn} recognized for their capabilities for time series analysis. Convolutional Neural Networks, which were primarily used for computer vision tasks, are now accepted for TSC tasks due to their capacity for extracting meaningful features in sequential data. In the work by \cite{Wang2017} architectures composed by CNNs could achieve state-of-the-art performance for TSC tasks, specifically a deep Residual Network (ResNet) and the Fully Convolutional Neural Network (FCN).

Among the most successful methods used for TC are LSTMs and CNNs. Despite their robustness, NNs do not give any quantifiable confidence in a classification problem. To address this, some UQ methods have now been integrated into Deep Learning (DL) approaches to make them trustworthy for critical decisions. The most commonly used UQ methodologies are Probabilistic Approximation and Ensemble Learning \cite{Abdar2021}. In our work, we focus on probabilistic approximation specifically Monte Carlo Dropout \cite{gal2016dropout}, DropConnect \cite{mobiny2021dropconnect} and Flipout \cite{wen2018flipout}. These Bayesian UQ techniques are increasingly being applied to DL to address the lack of quantifiable uncertainty and combine regularization methodologies. Ensembles are common but computationally demanding. We use Bayesian NNs which are scalable, easy to train and produce high-quality uncertainty.

% An early implementation of UQ is the construction of a neural network that can learn to represent probability distributions without receiving any direct probabilistic representations from the outside world \cite{Kharratzadeh2015}. As a NN is composed of deterministic units, this is done merely by observing the occurrence of patterns in events through probability matching.

\begin{figure}[h!]
\centering
\includegraphics[width=.5\textwidth]{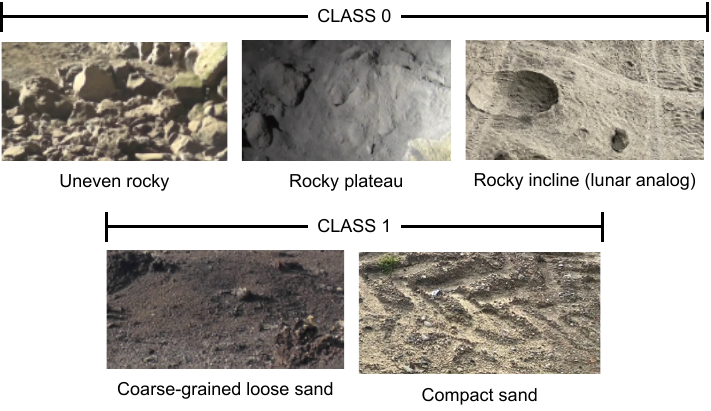}
\caption[Terrains]{The experimental sites for recording data include terrain that is mostly comprised of sand and rock. Rock is represented as uneven unconsolidated rock and flat rock plateaus landscapes or inclines shown as $Class_{0}$ in the first three images. Sand is represented present in a loose and compact form shown as $Class_{1}$ in the last two images. These images serve as visual aids for labeling and are not used to train our models.}
\label{fig:terrains}
\end{figure}

\section{Methods}
\label{sec:methods}

\begin{table*}[t]
\caption{Configuration space of networks. Values shown are ranges for the BO to select.}
\label{tab:conf_space}
\vskip 0.15in
\centering
\begin{tabular}{c|c|c|c|c|c|c|c}
  \textbf{CNN \& LSTM Layers} & \textbf{Filters} & \textbf{Kernels} & \textbf{Max Pooling} & \textbf{LSTM Cells} & \textbf{Batch Size} & \textbf{Dropout \%}\\
  \hline
  [1-3] & [16-128] & [4-16] & [2-8] & [8-128] & [16-64] & [0 - 0.5]\\
\end{tabular}
\vskip -0.1in
\end{table*}

% *********** Experimental platform ***********
Our experimental platform the AsguardIV, shown in Figure \ref{fig:asguard}, is a hybrid leg-wheel rover designed to allow navigation in unstructured environments. The rimless wheels are less complex, energy efficient and less prone to failures than articulated legs as they can adapt better when traversing obstacles and uneven terrain. This rover has been used to collect an array of data logs from trials executed across various locations with soil signatures representative of a lunar surface. The experimental sites are comprised of compact and loose sand, and rocky areas. For illustration purposes, we show in  Figure \ref{fig:terrains} the soil characteristics and level of traversal difficulty of compact and loose sand and rocky terrain from such sites. These images also aid in the labeling process but are not used to train our networks.

% *********** Datasets ***********

Binary classification for this task provides a clear contrast between two distinct types of terrain that would be found in a lunar landscape. Based on the characteristics of deformability and evenness, we label rock terrains as $Class_{0}$ for being uneven and undeformable, and sand terrains as $Class_{1}$, for even and deformable. 

Prior to formatting the inputs for the NNs, we remove any \textit{idle} gaps by trimming the streams. Here, \textit{idle} refers to prolonged gaps when the rover is not making progress. We explore two distinct settings for sequence generation to determine the optimal format. The first approach involves employing a sliding window with predefined width and step length, while the second approach entails sequence subsampling using a specified factor. For the sliding window sequence generation, we define a set $(W,S) = \{(100, 25), (400,100), (1000,100)\}$ of three pairs with different window size $w$ and step size $s$ in time steps.

In this approach, we extract consecutive and overlapping subsequences from the original time series by sliding a fixed-length window of size $w$ along the series with a step size of $s$. 
During sequence subsampling, we employ three different subsampling factors $f$ from the set $F = \{8, 16, 32\}$. These factors are selected heuristically to strike a balance between the number of samples and sample length. In this approach, we generate sequences by selecting every $f$-th time step from the original time series.

We evaluate all architectures with three different proprioceptive input configurations, the IMU and joint data independently and the fused IMU-joint data. The IMU array is a $6$-feature vector of the accelerometer and the gyroscope in three axes. The joint data is a time-synced $12$-feature vector of speed, acceleration, and effort for $4$ wheels. The fused data makes an input of an $18$-feature vector. The sensors are recorded at 100 Hz and compose a total of approximately 6 hours of data. We have split the data logs into  $70\%$ for training and $30\%$ for testing, ensuring that there is no correlation between test and train samples. The training set is further split into $80\%-20\%$ for training and validation.

We choose NNs for TC based on the work by \cite{Wang2017}. In this work, architectures composed by CNNs achieve state-of-the-art performance for TSC tasks, specifically a deep Residual Network (ResNet), the Fully Convolutional Neural Network (FCN), and an Attention Encoder. In addition, we design and tune our own LSTMs and CNNs.

% LSTMs are powerful at capturing dependencies in sequential data by solving the exploding and vanishing gradients problem \cite{Bengio1994}. This is the aftermath of continuous multiplication operations to calculate gradients which is present in the standard RNN. While CNNs are traditionally used for computer vision applications due to their ability to find spatial correlations from a 2-D or 3-D input image, 1-D convolutions can also be used for sequences \cite{Kiranyaz2021} by defining a sequence as a two-dimensional array of features across time. 

% The convolution process reduces the number of parameters and introduces sparsity to generate reduced feature vectors functionally performed by pooling the values and transforming them into a single output. In this manner, a CNN can support by finding the most relevant features in a time series.

% Hence, 1-D or 2-D convolutions can be employed by sliding the filters along the time dimension. 
% Unlike other NNs, CNNs share weights through the convolution operation. 
% that the LSTM takes in as input. 

We use a combined hyperparameter search and tuning using the automatic method Bayesian Optimization on Hyperband (BOHB) \cite{Falkner2018}. This allows us to speed up the tuning of our networks and generate only good-performing CNN and LSTM networks by combining state-of-the-art performance BO \cite{Bergstra2013,Snoek2015} with efficient resource allocation of HB \cite{Li2017}. The implementation for BOHB is available at: 
\url{https://github.com/automl/HpBandSter}.

% BOHB combines the efficiency of Bayesian Optimization (BO) with Hyperband \cite{Li2017} (HB). BO has already been successful in optimizing NNs to obtain state-of-the-art performance \cite{Bergstra2013,Snoek2015}. In BOHB, BO takes place by using an inexpensive fidelity of the objective function, a Tree-Parzen Estimator (TPE), optimizing in linear instead of cubic time. In addition, Hyperband dynamically allocates resources to the generated configurations and stops poorly performing configurations. The implementation for BOHB is available at: \url{https://github.com/automl/HpBandSter}.

% We generate three core architectures with BOHB given a set of parameter ranges which we have listed in Table \ref{tab:conf_space}. The first is a 1-D CNN that ranges from one to three CNN blocks. These blocks are composed of a 1-D CNN layer, a batch normalization layer, and a max-pooling layer. The Bayesian optimizer selects the number of blocks, number of filters, and kernel sizes in each block and maximum pooling size. Similarly. the second network is an LSTM that ranges from one to three layers. The Bayesian optimizer selects the number of cells. The third architecture is a CNN-LSTM where we connect $n$ CNN layers and $n$ LSTM layers. 

We generate three core architectures with BOHB namely CNN, LSTM and CNN-LSTM. The Bayesian optimizer selects the number of blocks, number of filters, and kernel sizes in each block, maximum pooling size for CNNs and number of layers and cells for LSTMs. General settings like batch size and Dropout percentage are also tuned. The parameter ranges are listed in Table \ref{tab:conf_space}. We use the TSC networks with the hyperparameters set according to \cite{Wang2017} and thus are not optimized with BOHB. After this process, we choose a set of candidate networks. 

Even with high-performing networks, their reliability is questionable in real-world applications. We address this by using three UQ techniques, namely Monte Carlo Dropout \cite{gal2016dropout}, DropConnect \cite{mobiny2021dropconnect} and Flipout  \cite{wen2018flipout}. Monte Carlo Dropout (MC Dropout) \cite{gal2016dropout} drops activations at inference time with probability $p$, while Monte Carlo DropConnect \cite{mobiny2021dropconnect} drops weights at a similar probability at inference time. Both methods produce an approximation to the predictive posterior distribution, which can be reconstructed through sampling. 

Flipout \cite{wen2018flipout} uses variational inference to approximate the posterior distribution for each weight with a Gaussian distribution, by maximizing the Evidence Lower Bound (ELBO). Flipout in particular is a variation of Bayes by Backprop \cite{Blundell2015}, where different kernel and bias samples are applied to each element in a batch, reducing variance during the training process and improving convergence. To make predictions, we estimate the predictive posterior distribution using $M = 10$ samples with Equation \ref{eq:ppd}, 

\newcommand{\condprob}[2]{
    \mathbb{P}(#1 \, | \, #2)
}

\begin{equation}
    \condprob{\mathbf{y}}{\mathbf{x}} \sim M^{-1} \sum_i^M \condprob{\mathbf{y}}{\theta_i, \, \mathbf{x}} \quad \theta_i \sim \condprob{\mathbf{w}}{\mathbf{x}}
    \label{eq:ppd}
\end{equation}

For Flipout, there is weight sampling, while in the case of MC Dropout and MC DropConnect, we only make forward passes $\condprob{\mathbf{y}}{\theta, \, \mathbf{x}}$. Monte Carlo Dropout layers are introduced after each 1-D Convolutional layer and before batch normalization in only the first two Convolutional blocks. This also applies if the architecture generated has only one CNN block. An MC Dropout layer is also applied before the classification layer if the architecture has LSTM layers. DropConnect Convolutional layers replace the standard layers in CNN-LSTM and CNN architectures. DropConnect Dense layers only replace the standard Dense classification layers in LSTM architectures. Flipout Dense layers are applied only to the output classification layer in all three architectures. 

\begin{figure*}
\centering
    \begin{subfigure}[b]{0.32\textwidth}
    \centering
    \includegraphics[width=\textwidth]{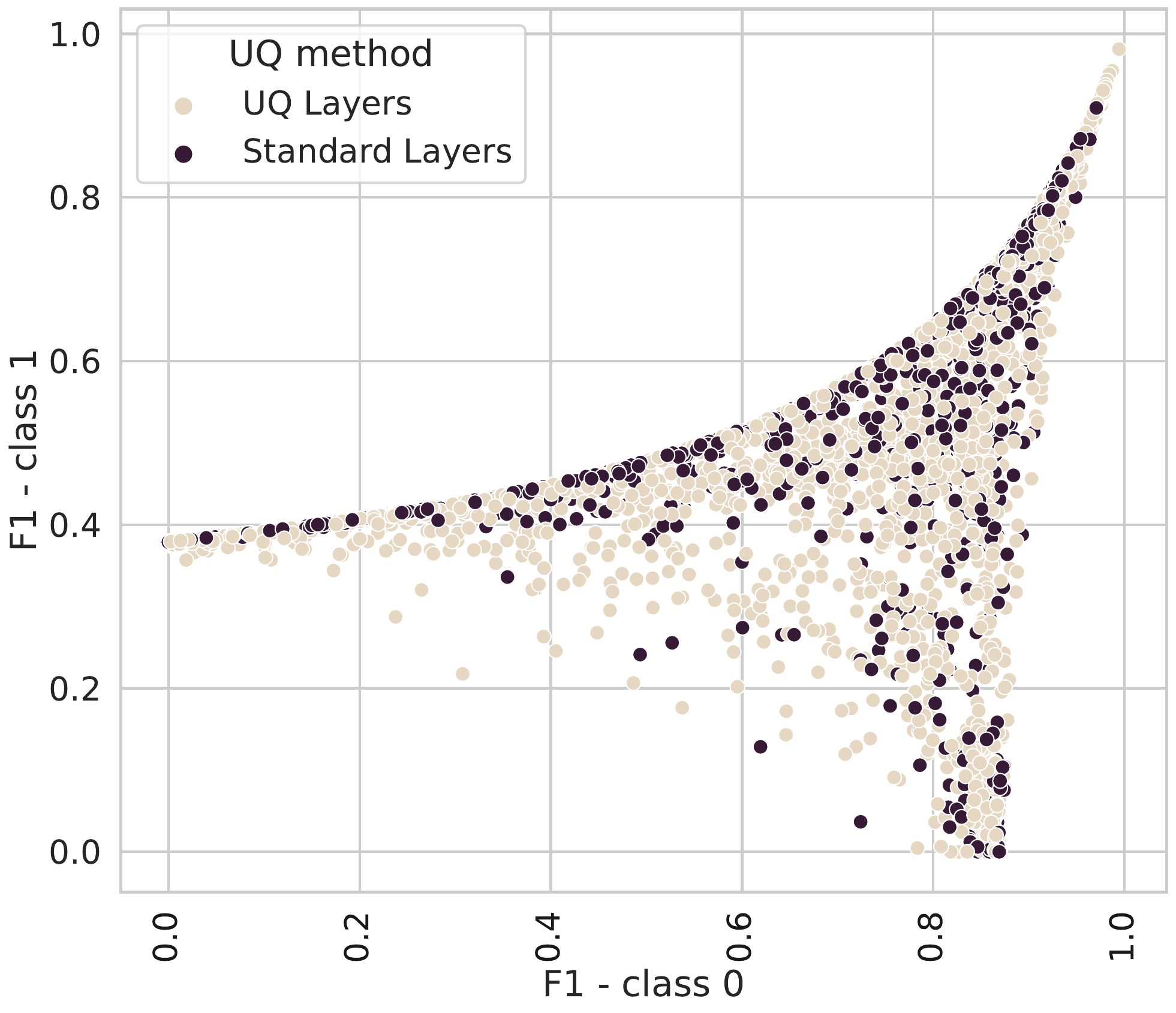}
    \caption{F1 score for each class}
    \label{fig:f1_cl0_cl1}
 \end{subfigure}
 \hfill
 \begin{subfigure}[b]{0.32\textwidth}
    \centering
    \includegraphics[width=\textwidth]{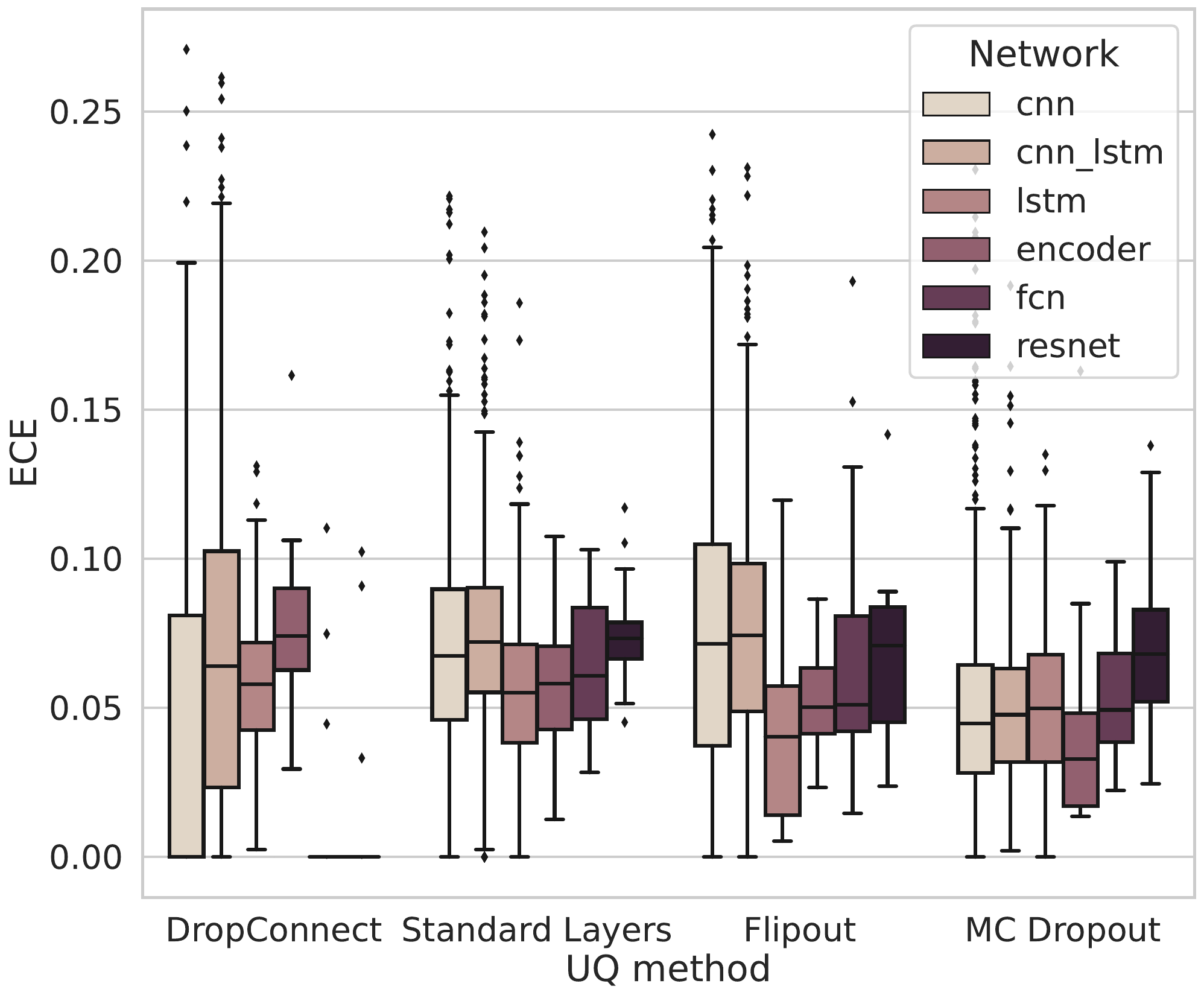}
    \caption{ECE per UQ method}
    \label{fig:ece_uq}
 \end{subfigure}
 \hfill
 \begin{subfigure}[b]{0.32\textwidth}
    \centering
    \includegraphics[width=\textwidth]{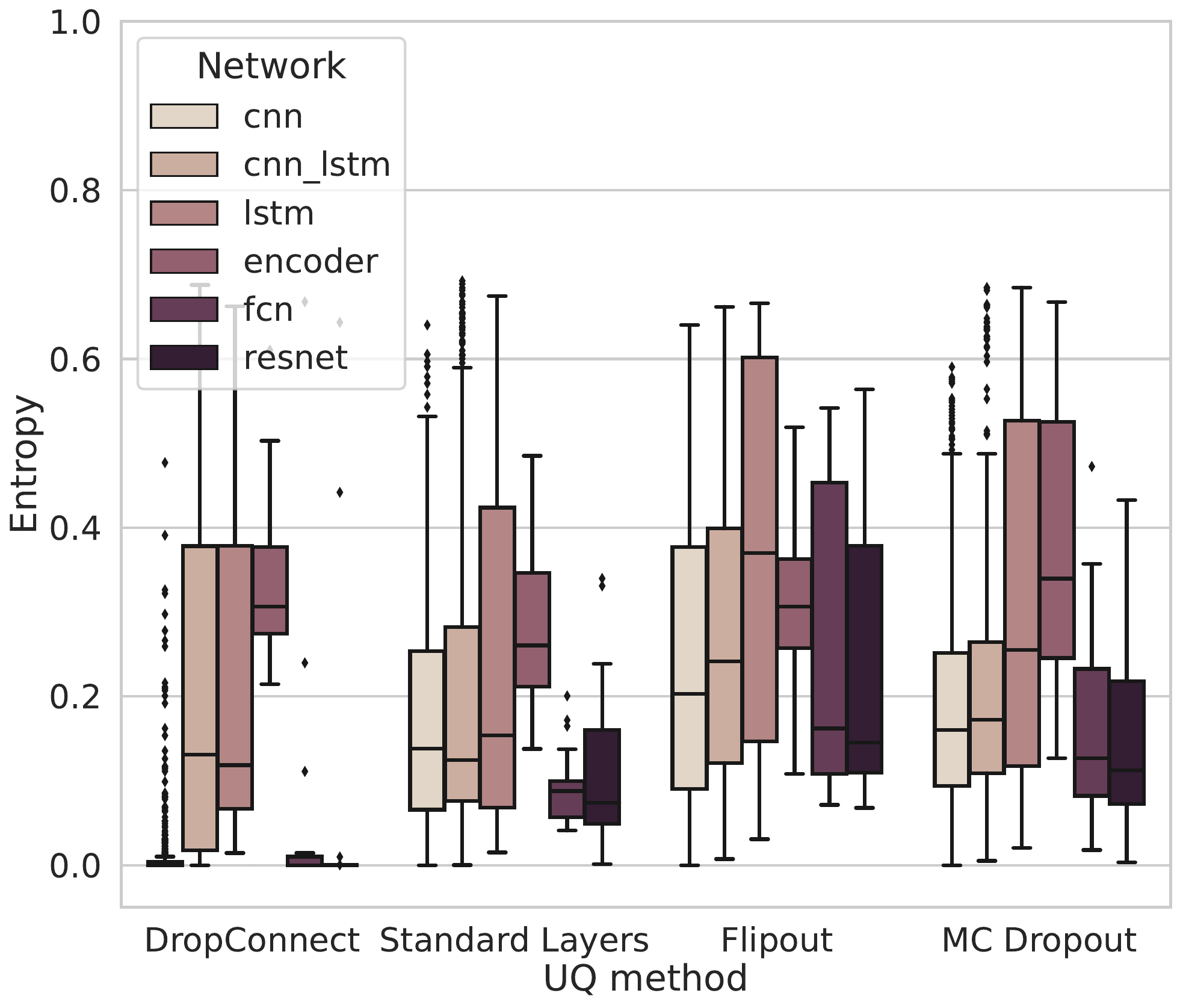}
    \caption{Entropy per UQ method}
    \label{fig:pe_uq}
 \end{subfigure}
 \hfill
\caption[All results UQ]{All candidates, both BOHB and TSC. (a) Performance behavior is observed by comparing $F1_{cl_{0}}$ vs. $F1_{cl_{1}}$. (b) ECE scores by UQ and architecture. (c) Predictive entropy scores by UQ and architecture.}
\label{fig:all_res_bohb_dl4tsc}
\end{figure*}

\begin{figure*}[ht]
	\centering
	\begin{subfigure}[b]{0.24\textwidth}
		\centering
		\includegraphics[width=\textwidth]{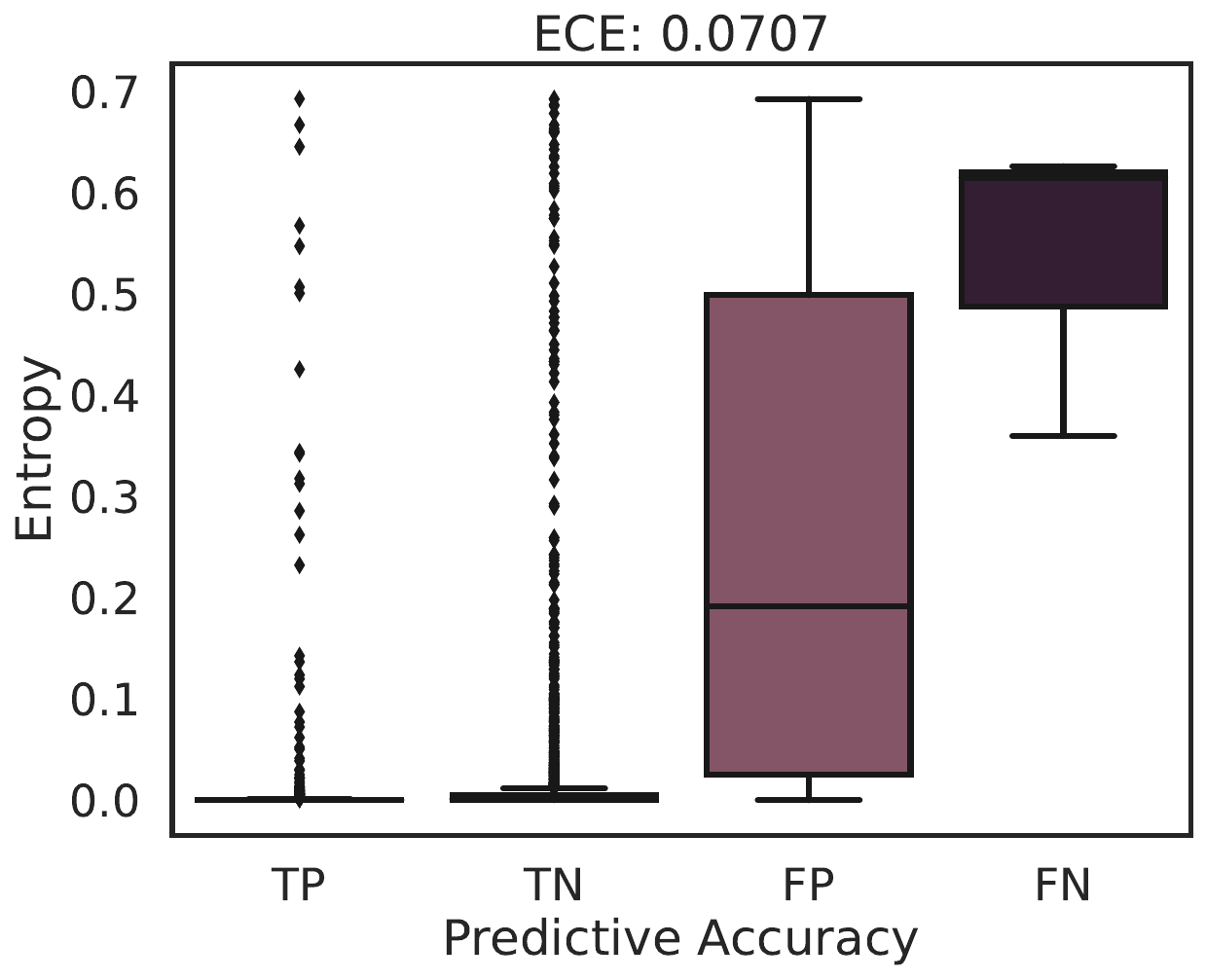}
		\caption{LSTM - DropConnect}
		\label{fig:dropconnect}
	\end{subfigure}
	% \hfill
	\begin{subfigure}[b]{0.24\textwidth}
		\centering
		\includegraphics[width=\textwidth]{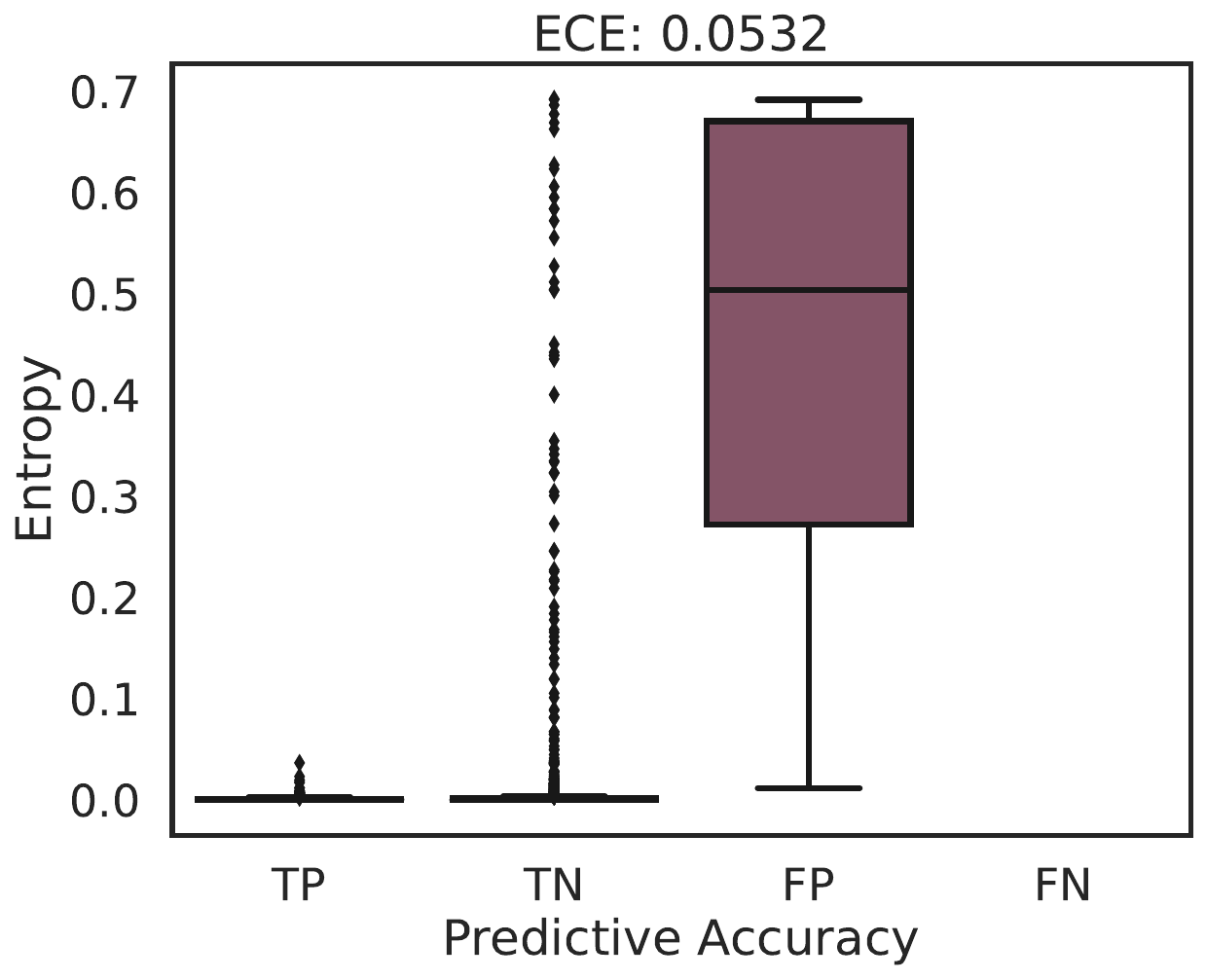}
		\caption{CNN-LSTM - MC Dropout}
		\label{fig:dropout}
	\end{subfigure}
	\hfill
	\begin{subfigure}[b]{0.24\textwidth}
		\centering
		\includegraphics[width=\textwidth]{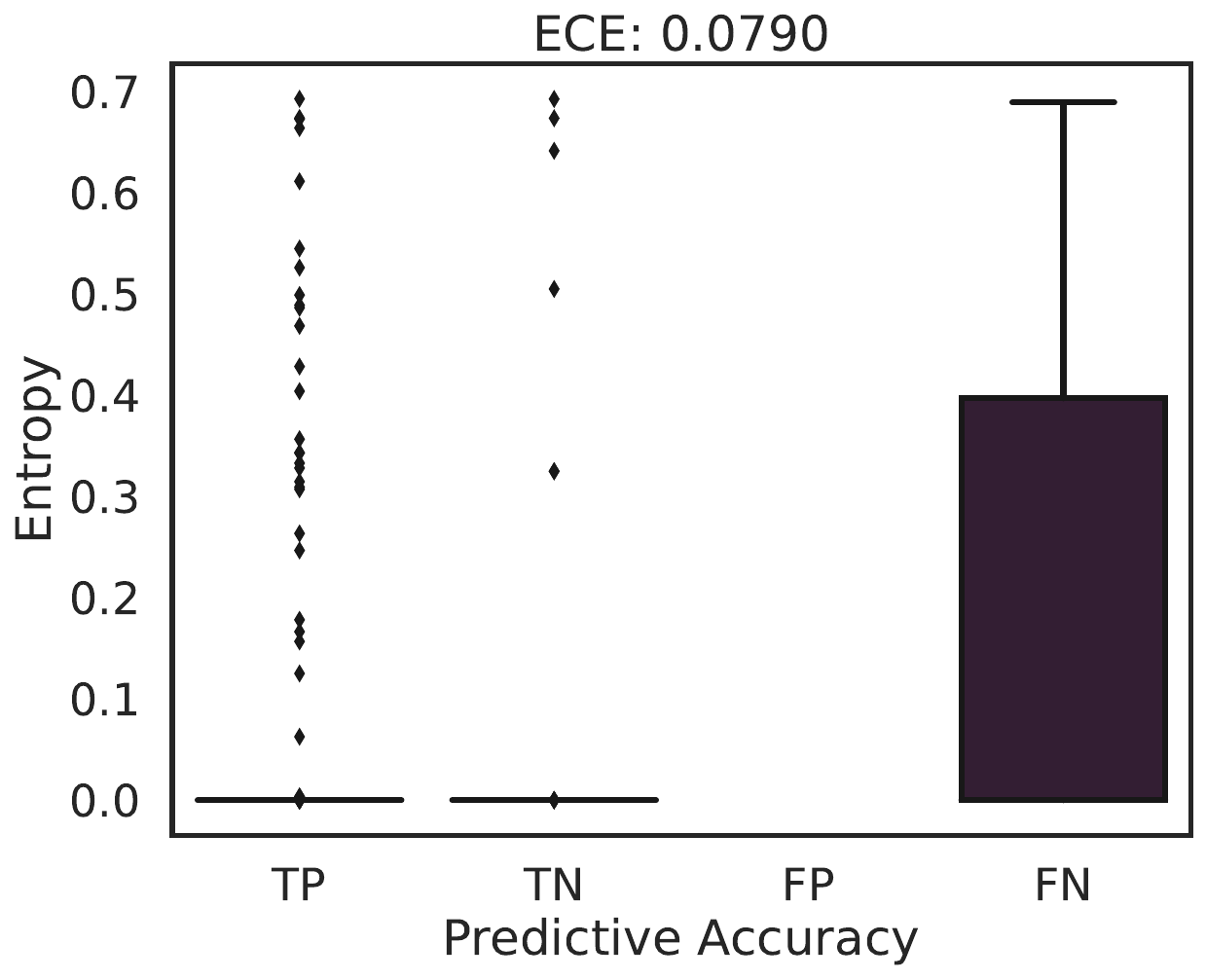}
		\caption{CNN - Flipout}
		\label{fig:flipout}
	\end{subfigure}
	% \hfill
	\begin{subfigure}[b]{0.24\textwidth}
		\centering
		\includegraphics[width=\textwidth]{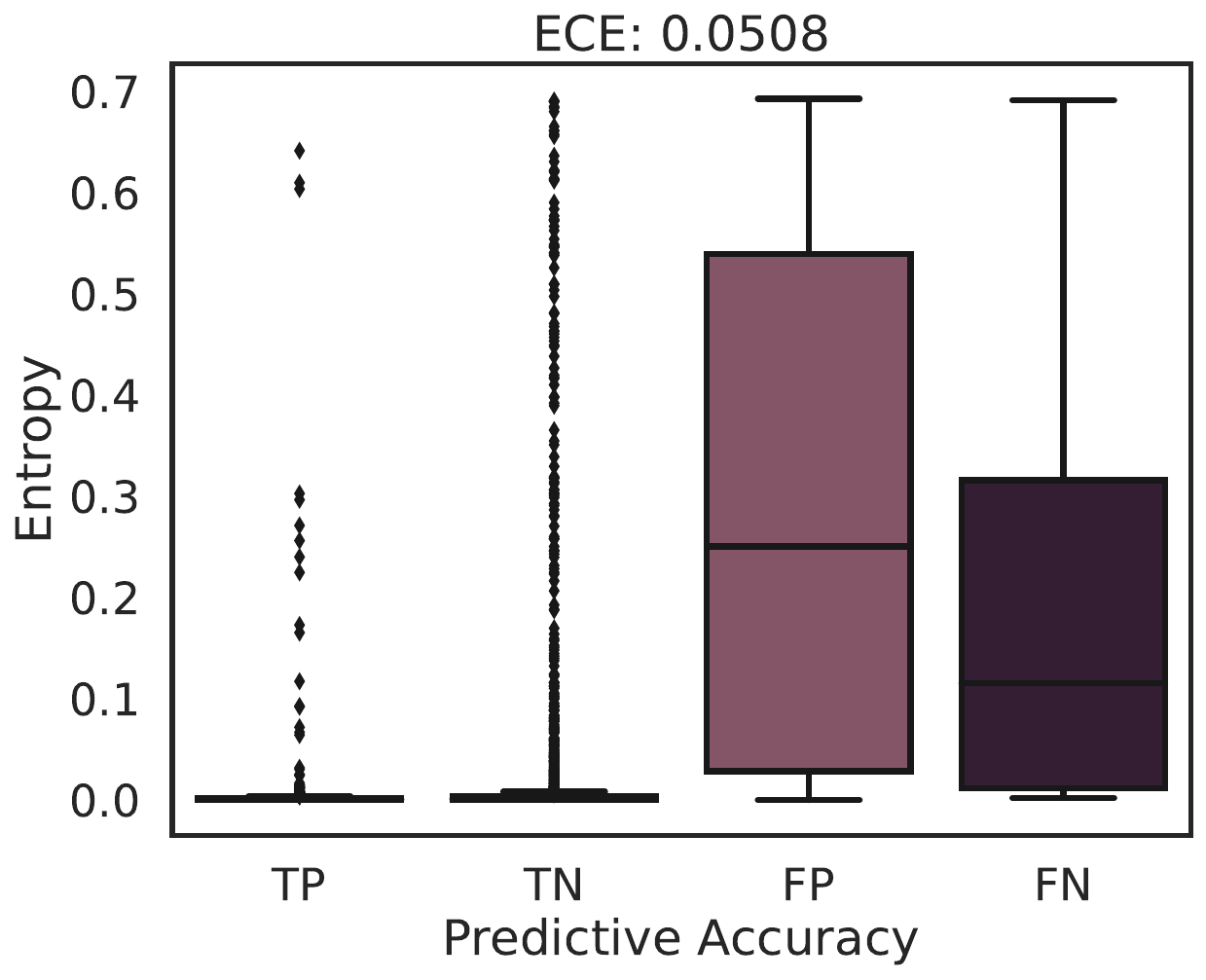}
		\caption{LSTM - Standard Layer}
		\label{fig:none}
	\end{subfigure}
	\hfill
\caption[PredAcc v Entropy]{Predictive accuracy versus entropy from the selected candidates for each UQ method.} %in Table \ref{tab:best_params}, i.e. selected candidate per UQ method.
\label{fig:predacc_v_entropy}
\end{figure*}

Similarly, the TSC architectures, ResNet, FCN and Attention Encoder from \cite{Wang2017}, are also enhanced with UQ layers. DropConnect layers substitute the traditional ones in the ResNet only for the convolutional blocks and not the \textit{shortcut} block used for channel expansion. All layers are replaced for the FCN architecture and the Attention Encoder. MC Dropout and Flipout Dense layers replace all standard Dropout layers and Dense classification output layers respectively. 

We assess all UQ networks against their non-UQ counterparts. The assessment criteria that we use for UQ is predictive entropy which measures the uncertainty of the prediction. This is useful to detect classifications from high-performing networks with high uncertainty. This metric is given by Equation \ref{eq:entropy}, 
% $ H(\mathbb{P}) = -\sum_{c \in C} \mathbb{P}(c) \log \mathbb{P}(c)$, with $c \in C$ the set of classes, and $\mathbb{P}(c)$ the probability of class $c$.

\begin{equation}
    H(\mathbb{P}) = -\sum_{c \in C} \mathbb{P}(c) \log \mathbb{P}(c)
    \label{eq:entropy}
\end{equation}

with $c \in C$ the set of classes, and $\mathbb{P}(c)$ the probability of class $c$. In addition, we use ECE as a metric to assess the calibration of our candidate models. By quantifying the difference between predicted probabilities and the actual frequencies, ECE provides a measure of confidence and reliability in the predictive outputs of the model. This score is calculated by Equation \ref{eq:ece} to calculate the empirical probabilities of all instances that fall into bin $i$,

\begin{equation}
    ECE = \sum_{i=1}^{K} P(i) \cdot |o_{i} - e_{i}|
    \label{eq:ece}
\end{equation}

where $o_{i}$ is the true fraction of positive instances and $e_{i}$ the mean of post-calibrated probabilities.

\begin{table*}[t]
\centering
\caption{Performance and hyperparameters of selected networks with UQ methods of MC Dropout (DO), DropConnect (DC), and Flipout (FO) and with Standard Layers (SL).}
\label{tab:best_params}
\vskip 0.15in
\resizebox{\textwidth}{!}{%
\begin{tabular}{c|l|c|c|c|c|c|c|c|c|c|c}
	ID & Study & UQ & Entropy & ECE & $F1_{cl_{0}}$ &  $F1_{cl_{1}}$ & $F1_{weighted}$ & Acc & Dropout rate & CNN & LSTM \\  
\hline
& CNN \\	% 21 models															
\hline
 1 &	jstate-imu-in1000-step100 & DO & 0.0739 & 0.0134 & 0.9726 & 0.9110 & 0.9582 & 0.9581 & 0.46 & $\{f_{1}:22, f_{2}:21, k_{1}:11, k_{2}:16\}$ \\
2 &	imu-in1000-step100 	     & SL & 0.0374 & 0.0556 & 0.9704 & 0.9095 & 0.9562 & 0.9554 & 0.21 & $\{f_{1}:26, f_{2}:19, k_{1}:6, k_{2}:7 \}$ \\
3 &	imu-in400-step100         & DO & 0.0843 & 0.0122 & 0.9787 & 0.9280 & 0.9669 &	0.9672 &	0.42 &	$\{f_{1}:16, k_{1}:7\}$ \\
4 &	imu-in400-step100     &	DO & 0.0451 & 0.0354 & 0.9801 &	0.9342 & 0.9694 & 0.9695 &	0.32 &      $\{f_{1}:77, k_{1}:11\}$ \\
5 &	imu-in400-step100        &	DO & 0.0410  & 0.0493 & 0.9704 &	0.9050 & 0.9551 & 0.9548 &	0.34 &	$\{f_{1}:77, f_{2}:22, k_{1}:14, k_{2}:8 \}$ \\
6 &	imu-in1000-step100 	 	& DO & 0.0707 & 0.0122 & 0.9742 &  0.9171 &	0.9608 &	0.9606 	&0.50 &	$\{f_{1}:16, k_{1}:16 \}$\\
7 &	imu-in1000-step100 	& DO &	0.0381 & 0.0355 &	0.9701 & 	0.9120 &	0.9565 &	0.9554 &	0.18 &	$\{f_{1}:16, k_{1}:6 \}$ \\
8 &	imu-in1000-step100 & DO &	0.0211 & 0.0621 &	0.9813 &	0.9430 &	0.9724 &	0.9719 	& 0.11 &	$\{f_{1}:52, k_{1}:13 \}$ \\
9 &	imu-in1000-step100 & DO &	0.0405 & 0.0290 &	0.9678 &  0.9071 &	0.9536 &	0.9522 &	0.11 &	$\{f_{1}:54, k_{1}:9 \}$ \\
10 &	imu-in1000-step100 & DO &  0.0693 &  0.0264 & 0.9723 &	0.9177 &	0.9596 &	0.9586 & 0.43 	& $\{f_{1}:63, f_{2}:41, k_{1}:8, k_{2}:7\}$\\
11 &	imu-in1000-step100 & DO & 0.0433 & 0.0238  &	0.9761 & 	0.9255 &   0.9643 & 0.9638 	&0.15 & $\{f_{1}:47, k_{1}:10\}$ \\
12 &	imu-in1000-step100	& DO &	0.0381 & 0.0291 &	0.9798 &	0.9375 &	0.9699 &	0.9694 	&0.23 &	$\{f_{1}:56, f_{2}:78, k_{1}:7, k_{2}:4\}$ \\
13 &	imu-in1000-step100  & DO &	0.0402 &  0.0127 & 0.9800  &	0.9388 & 0.9704 &	0.9698 &	0.09 &	$\{f_{1}:19, k_{1}:14 \}$ \\
14 &	imu-in1000-step100 & DO	& 0.0187 & 0.0494 &	0.9840 &	0.9508 &	0.9763 &	0.9759 &	0.02 &	$\{f_{1}:29, k_{1}:11 \}$ \\
15 &	imu-in1000-step100 & DO &	0.0323 & 0.0644 &	0.9730 &	0.9181 &	0.9602 &	0.9594 &	0.23 &	$\{f_{1}:51, f_{2}:75, f_{3}:57, k_{1}:11, k_{2}:6, k_{3}:6 \}$ \\
16 &	imu-in400-step100 &	FO &	0.0710 & 0.0256 &	0.9797 	& 0.9315 &	0.9685 &	0.9687 &	0.40 &	$\{f_{1}:97, k_{1}:15 \}$ \\
17 &	imu-in400-step100 &	FO &	0.0261 & 0.0515 &	0.9761 &	0.9132 &	0.9614 &	0.9625 &	0.01 &	$\{f_{1}:116, k_{1}:14 \}$ \\
18 &	\textbf{imu-in400-step100} &	\textbf{FO} &  \textbf{0.0096} & \textbf{0.0790} &	\textbf{0.9871} &	\textbf{0.9550} &	\textbf{0.9796} &	\textbf{0.9799} &	0.17 &  $\{f_{1}:16, f_{2}:63, f_{3}: 102,  k_{1}:6, k_{2}:5, k_{3}:11\}$ \\
19 &	imu-in1000-step100 & FO &	0.0413 & 0.0296 &	0.9784 &	0.9336 &	0.9679 &	0.9674 &	0.10 &  $\{f_{1}:23, k_{1}:6\}$ \\
20 &	imu-in1000-step100 & FO & 	0.0227 & 0.0343 &	0.9843 &	0.9485 &	0.9759 &	0.9759 &	0.02 &  $\{f_{1}:46, f_{2}:36, f_{3}:17, k_{1}:12, k_{2}:14, k_{3}:14\}$ \\
21 &	imu-in1000-step100 & DO &	0.0252 & 0.0369 &	0.9819 &	0.9445 &	0.9731 &	0.9727 &	0.17 &	$\{f_{1}:107, f_{2}:28, k_{1}:12, k_{2}:5\}$ \\ 

\hline
& CNN-LSTM  \\  % 6 models
\hline
22 &	imu-in400-step100 &	DO &	0.0629 & 0.0235 &	0.9775 &	0.9308 &	0.9666 	& 0.9660 	& 0.06 & $\{f_{1}:64, f_{2}:62, k_{1}:12, k_{2}:9\}$ & $\{u_{1}:8\}$ \\ 
23 &	imu-in1000-step100 & DO &	0.0163 & 0.0556 & 0.9670 &	0.9044 &	0.9524 &	0.9509 &	0.02 & $\{f_{1}:53, k_{1}:12 \}$ & $\{u_{1}:22, u_{2}:18, u_{3}:36\}$ \\
24 &	\textbf{imu-in1000-step100} & \textbf{DO} &	\textbf{0.0142} & \textbf{0.0532} &	\textbf{0.9942} &	\textbf{0.9814} &	\textbf{0.9912} &	\textbf{0.9912} &	0.04 & $\{f_{1}:26, k_{1}:9 \}$ & $\{u_{1}:67 \}$ \\
25 &	imu-in1000-step100  & DO &	0.0497 &  0.0152 &  0.9705 & 	0.9120 &	0.9568 &	0.9558 &	0.12 & $\{f_{1}:54, k_{1}:12\}$ & $\{u_{1}:11, u_{2}:59\}$ \\
26 &	imu-in1000-step100 &  FO &	0.0417 & 0.0444 &	0.9816 &	0.9362 &	0.9710 &	0.9715 &	0.01 & $\{f_{1}:79, f_{2}:94, f_{3}:105, k_{1}:8, k_{2}:14, k_{3}:4\}$ & $\{u_{1}:18, u_{2}:9\}$ \\
27 &	imu-in1000-step100 &  FO &	0.0675 & 0.0466 &	0.9718 &	0.9165 &	0.9588 &	0.9578 &	0.06 & $\{f_{1}:94, f_{2}:27, k_{1}:5, k_{2}:5\}$ & $\{u_{1}:65, u_{2}:10\}$ \\
\hline
& LSTM  \\ % 9 models
\hline
28 &	imu-in400-step100 &	SL &	0.0596 & 0.0483 &	0.9760 &	0.9253 &	0.9642 &	0.9637 &	0.0  & & $\{u_{1}:41\}$ \\
29 &	jstate-imu-in400-step100 &	DC &0.0347 & 0.0580 &	0.9689 &	0.9091 &	0.9549 &	0.9536 &0.25 & & $\{u_{1}:113, u_{2}:13 \}$ \\
30 &	imu-in1000-step100     & DC 	& 0.0304 & 0.0531 &	0.9791 	& 0.9337 &	0.9685 &	0.9682 &	0.25 & & $\{u_{1}:93\}$ \\
31 &	\textbf{imu-in400-step100} &	\textbf{DC} 	& \textbf{0.0361} & \textbf{0.0707} & \textbf{0.9664} 	& \textbf{0.9032} &	\textbf{0.9517} &	\textbf{0.9502} &	0.25 & & $\{u_{1}:23, u_{2}:74\}$ \\
32 &	imu-in400-step100 &	DC &	0.0383 & 0.0587 & 0.9690 &	0.9078 & 0.9547 &	0.9537 &	0.25 & & $\{u_{1}:53 \}$ \\
33 &	imu-in400-step100 &	DO &	0.0868 & 0.0616 &	0.9714 & 	0.9148 &	0.9581 &	0.9571 &	0.25 & & $\{u_{1}:24, u_{2}:10 \}$ \\
34 &	imu-in400-step100 &	DO &	0.0395 & 0.0656 &	0.9675 &	0.9028 &	0.9524 &	0.9513 &	0.25 & & $\{u_{1}:95 \}$ \\
35 &	imu-in400-step100 &	SL &	0.0607 & 0.0519 &	0.9708 &	0.9075 &	0.9560 &	0.9556 &	0.0 & & $\{u_{1}:44\}$ \\
36 &	\textbf{imu-in400-step100} &	\textbf{SL} &	\textbf{0.0367} & \textbf{0.0508} &	\textbf{0.9701} &	\textbf{0.9109} &	\textbf{0.9562} &	\textbf{0.9552} &	0.0  & & $\{u_{1}:53\}$ \\
\hline
& ResNet  \\ % 1 model
\hline
37 &	imu-in1000-step100 & FO &	0.0812 & 0.0238 &	0.9683 &	0.9085 &	0.9544 &	0.9530 &	0.25 & & \\	

% %%%%%%%%%%%%%%%%%%%%%%%%%%%%%%%%%%%%%%%%%%%%%%%%%%%%%%%%%%%%%%%%%%%%%%%%%%%%%%%
% %%%%%%%%%%%%%%%%%%%%%%%%%%%%%%%%%%%%%%%%%%%%%%%%%%%%%%%%%%%%%%%%%%%%%%%%%%%%%%%
\end{tabular}
}
\end{table*}

\section{Results and Discussion}
\label{sec:res}

We train all the networks with the Nvidia GeForce RTX 3070 and RTX 2070 graphics cards. We use Adam \cite{Kingma2014} as our optimization algorithm with an initial learning rate of $10^{-2}$. The maximum budget given to the Bayesian optimizer was $50$ epochs, successive halving being executed at the $16_{th}$ epoch.  We conduct a total of $216$ BOHB studies, each with different input data, UQ settings, and architecture search configurations. The studies were run for $20$ BO iterations resulting in a pool of $6,480$ full-budget candidates. Similarly, the experiments for TSC architectures result in a total of $216$ candidates.

We analyze the performance of UQ versus non-UQ networks paying attention to the $F1$ scores for each class, i.e. the harmonic mean of precision and recall. Sequences corresponding to $Class_{1}$ are underrepresented in our dataset. The percentage varies according to each sequence generation setting but it approximately averages to 50\%. Hence, we are interested in high $F1$ scores for both classes. The overall performance of all the networks is shown in Figure \ref{fig:all_res_bohb_dl4tsc}. We disregard deficient models and present the performance of $6,603$ candidates by comparing $F1$ scores for each class, ECE per architecture, and Entropy per architecture. 

Figure \ref{fig:f1_cl0_cl1} shows how BOHB ensures that many candidates reach optimal accuracy performance for both UQ-enabled and non-UQ architectures. However, it is clear from Figure \ref{fig:ece_uq} that MC Dropout architectures achieve better calibration scores whereas Figure \ref{fig:pe_uq} illustrates how DropConnect architectures appear to have even lower uncertainty than other UQ methods.  To assess this more closely, we select a set $S$ of candidates that meet the following criteria,

\begin{equation}
S =
\begin{cases}
\begin{aligned}
\text{Select}, & \text{ if } (F1_{cl_{0}} \land F1_{cl_{1}}) >= 0.9 \\
& \text{ and } H(\mathbb{P}) <= 0.1\\
\text{Reject}, & \text{ if } (F1_{cl_{0}} \lor F1_{cl_{1}}) < 0.9 \\
& \text{ or } H(\mathbb{P}) > 0.1
\end{aligned}\\
\end{cases}
\label{eq:sel_criteria}
\end{equation}

Table \ref{tab:best_params} shows the $37$ selected models with this criteria.  We identify the studies by the sensor input and some variation in sequence structure. We then list their Uncertainty Quantification method, performance by predictive entropy, ECE, $F1$ scores for both classes and weighted score, accuracy, and the hyperparameters for the candidates that have been tuned with BOHB. 

We observe that models trained only with IMU input significantly outperform models trained with other sensor inputs. Additionally, we find that sequences generated with the sliding window approach achieve better performance compared to subsampled sequences. This suggests that when dealing with distinct mobility signatures, subsampling may result in the loss of crucial information, leading to underperformance.

Establishing a direct impact of the input size on the performance of the model is an interesting task given the number of settings and parameters. Out of a total of $37$ selected models, 10-second inputs are required for $22$ of them, while the remaining $15$ require 4-second inputs. Although longer 10-second sequences may enhance the identification of characteristic terrain features, we observe that the performance of the shorter 4-second sequences is comparable to their 10-second counterparts. These results are important for evaluating the impact of input size on model performance and suggest that shorter input sequences may be sufficient to achieve robust classification in a planetary exploration scenario where computational advantages are necessary.

We highlight interesting values from the selected studies in bold. From this set, we choose one candidate with low entropy for each UQ method and analyze the performance as shown in Figure \ref{fig:predacc_v_entropy}. Here, we display four important details: ECE, Entropy, and Predictive Accuracy values, and architecture.

We note that each candidate has a good calibration score. While this varies on each application, it is generally accepted that values below $0.1$ are considered well-calibrated. In addition, we expect that incorrect predictions, FPs and FNs, should have higher entropy, i.e. be more uncertain, than correct predictions, TPs and TNs. This assumption holds again for the MC Dropout candidate with no misclassifications for \textit{$Class_{1}$}. Flipout has no FPs and wide uncertainty for FNs. Standard Layers and DropConnect are comparable in uncertainty performance and surpassed by MC Dropout. We also note that these candidates are all BOHB networks.

Overall, we believe that these results show the usefulness of UQ predictive entropy when used to assess the quality and likelihood of a prediction being correct.
For our Terrain Classification problem, NNs without uncertainty are inadequate and only MC Dropout models provide reliable classifications. TSC architectures appear to be less variable in UQ and calibration performance, but this is due to the number of total candidates compared to BOHB.

\section{Conclusion and Future Work}
\label{sec:conclusion}

We focus on the advantage of UQ in a problem of Terrain Classification for an exploration rover motivated by the need for reliable and trustworthy NNs. We compare a wide range of NN candidates using fast and efficient BOHB optimization and UQ methods. We have also compared our models with UQ-enabled benchmarks: ResNet, FCN and Attention Encoder. 

In the context of space missions, models that produce high-confidence outputs with low uncertainty can ensure navigation safety by enabling navigation with hazardous terrain avoidance. Our results demonstrate a clear advantage to integrating UQ into TC. This furthermore provides higher trustworthiness during planetary exploration. 

Our future work involves online testing our networks in analogous scenarios to provide insights into the limitations of binary TC and extend the classification to additional terrain types for terrestrial missions. We also aim to employ multiple-objective optimization techniques. By incorporating entropy as an additional performance metric, we seek to generate neural network configurations that optimize both high performance and low entropy simultaneously. This approach will enable us to further refine the training process and achieve more robust and balanced models.

\section*{Acknowledgements}

We gratefully acknowledge the support and funding provided by projects Insys and Persim funded by the Space Agency of the German Aerospace Center with federal funds of the Federal Ministry of Economic Affairs and Climate Action (BMWK) in accordance with the parliamentary resolution of the German Parliament, grants no. 50RA2036 and 50RA2124.

% \textbf{Do not} include acknowledgements in the initial version of
% the paper submitted for blind review.

% If a paper is accepted, the final camera-ready version can (and
% probably should) include acknowledgements. In this case, please
% place such acknowledgements in an unnumbered section at the
% end of the paper. Typically, this will include thanks to reviewers
% who gave useful comments, to colleagues who contributed to the ideas,
% and to funding agencies and corporate sponsors that provided financial


% \clearpage
\balance
\bibliography{terrain_clf}
\bibliographystyle{icml2023}

%%%%%%%%%%%%%%%%%%%%%%%%%%%%%%%%%%%%%%%%%%%%%%%%%%%%%%%%%%%%%%%%%%%%%%%%%%%%%%%
%%%%%%%%%%%%%%%%%%%%%%%%%%%%%%%%%%%%%%%%%%%%%%%%%%%%%%%%%%%%%%%%%%%%%%%%%%%%%%%
% APPENDIX
%%%%%%%%%%%%%%%%%%%%%%%%%%%%%%%%%%%%%%%%%%%%%%%%%%%%%%%%%%%%%%%%%%%%%%%%%%%%%%%
%%%%%%%%%%%%%%%%%%%%%%%%%%%%%%%%%%%%%%%%%%%%%%%%%%%%%%%%%%%%%%%%%%%%%%%%%%%%%%%
% \newpage
% \appendix
% \onecolumn
% \section{Appendix.}

%If you want, you can use an appendix like this one, even using the one-column format.

\end{document}